\documentclass{article}

\usepackage[numbers]{natbib}
\usepackage[preprint,nonatbib]{neurips_2024}

\usepackage[utf8]{inputenc} 
\usepackage[T1]{fontenc}    
\usepackage{hyperref}       
\usepackage{url}            
\usepackage{booktabs}       
\usepackage{amsfonts}       
\usepackage{nicefrac}       
\usepackage{microtype}      
\usepackage{xcolor}         
\usepackage{graphicx}
\usepackage{amsmath}
\usepackage{amsthm}
\usepackage{amssymb}
\usepackage{subcaption}
\usepackage{booktabs}
\usepackage{xcolor}
\usepackage{multirow}
\usepackage{wrapfig}
\usepackage{algorithm}
\usepackage{algorithmic}

\newtheorem{theorem}{\textbf{Theorem}}

\usepackage{titlesec}
\usepackage{setspace}
\title{Inaccurate Label Distribution Learning with Dependency Noise}

\author{%
  Zhiqiang Kou, Jing Wang, Yuheng Jia, and Xin Geng\thanks{correspondingauthor} \\
  MOE Key Laboratory of Computer Network and Information Integration, \\
  School of Computer Science and Engineering, Southeast University, Nanjing, China\\
  \{zhiqiang\_kou, wangjing91, yhjia, xgeng\}@seu.edu.com
}

\begin{document}

\maketitle

\begin{abstract}
In this paper, we introduce the Dependent Noise-based Inaccurate Label Distribution Learning (DN-ILDL) framework to tackle the challenges posed by noise in label distribution learning, which arise from dependencies on instances and labels. We start by modeling the inaccurate label distribution matrix as a combination of the true label distribution and a noise matrix influenced by specific instances and labels. To address this, we develop a linear mapping from instances to their true label distributions, incorporating label correlations, and decompose the noise matrix using feature and label representations, applying group sparsity constraints to accurately capture the noise. Furthermore, we employ graph regularization to align the topological structures of the input and output spaces, ensuring accurate reconstruction of the true label distribution matrix. Utilizing the Alternating Direction Method of Multipliers (ADMM) for efficient optimization, we validate our method's capability to recover true labels accurately and establish a generalization error bound. Extensive experiments demonstrate that DN-ILDL effectively addresses the ILDL problem and outperforms existing LDL methods.

\end{abstract}

\section{Introduction}

Label Distribution Learning (LDL) \cite{geng2016label, jia2021label} is an innovative learning approach where each instance is associated with a label distribution. Fundamentally, a label distribution is a multi-dimensional vector, with elements known as label description degrees that signify the relative significance of each label \cite{wang2021label}.  Fig. 1 presents an image from a natural-scene dataset \cite{geng2014multilabel}. The average ratings have been adjusted to create a label distribution $\{0.25, 0.4, 0.3, 0.05\}$, which represents the varying levels of significance attributed to each label. scholars.\cite{xu2017incomplete}. LDL explicitly trains a model to associate instances with label distributions. In contrast to single-label learning (SLL) and multi-label learning (MLL), LDL directly addresses label ambiguity, garnering significant interest among Label distribution learning methods typically rely on precise label information in training data.\begin{wrapfigure}[8]{r}{0.4\textwidth}
	\centering
	\includegraphics[width=0.4\textwidth]{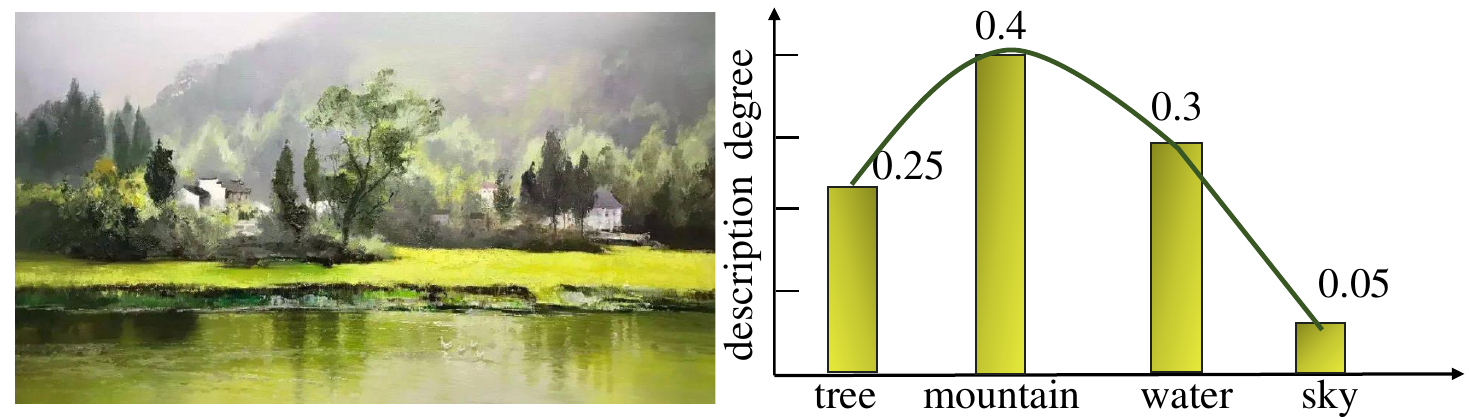}
	\caption{An image from a natural-scene dataset \cite{geng2014multilabel} with a label distribution. }
	\label{fig1}
\end{wrapfigure} However, creating extensive and high-quality labeled datasets poses significant challenges, primarily due to the frequent occurrence of inaccurate annotations. For instance, in tasks like movie sentiment analysis, annotators are assigned to label movie reviews based on emotions, such as positive or negative sentiments. Given the subjective nature of sentiment analysis, an annotator might mistakenly classify a review expressing happiness as one conveying surprise. Such subjective errors introduce inaccuracies in the labeled dataset, potentially affecting the performance of LDL models.

These inaccuracies often manifest as noisy labels within the training set. In response, a novel framework called \textit{"Inaccurate Label Distribution Learning"} has emerged and attracted attention \cite{kou2023instance}. Kou \cite{kou2023inaccurate} first introduced this concept, where the label distribution matrix is perturbed by random noise, such as Gaussian noise, salt-and-pepper noise, or Laplacian noise. They proposed a two-stage approach to recover the ideal label distribution from the noisy label distribution and train the classifier.  Next, LRS-LDL \cite{kou2023instance} addresses the problem of instance-dependent inaccurate label learning, where noisy labels are associated with instances. They proposed a classifier learning framework based on inaccurate label distribution.

Existing algorithms often assume that label noise is either independent of both labels and instances or solely dependent on instances. However, these assumptions may not hold in practical scenarios. Firstly, the likelihood of noisy labeling can vary across different class labels, a phenomenon known as \textit{label-dependent label noise}. For instance, in an image classification dataset distinguishing "cat" from "dog," the label "dog" might be more susceptible to noise due to visual similarities with other canines such as wolves or foxes. This variability in noise susceptibility highlights the presence of label-dependent label noise. Secondly, even within the same label category, instances may exhibit vastly different feature representations, influencing their propensity for mislabeling. This is referred to as \textit{instance-dependent label noise}. Consider a sentiment analysis task where text snippets are categorized as "positive" or "negative." Two snippets both labeled as "positive"—one describing joy about a sunny day and another celebrating a game victory—may have different risks of mislabeling. The subtle language of the first snippet might render it more prone to being incorrectly labeled as "neutral," compared to the more straightforward second snippet. Thus, both \textit{instance-dependent} and \textit{label-dependent} label noises are significant factors in real-world scenarios, yet they remain underexplored in existing research. 

In this paper, we propose the Dependent Noise-based Inaccurate Label Distribution Learning (DN-ILDL) method to tackle the issue of dependent noise. We begin by modeling the inaccurate label distribution matrix as a combination of the true label distribution and a dependent noise matrix. We then develop a linear mapping \cite{rassias1978stability} from instances to their true label distributions, taking into account label correlations \cite{xu2017incomplete}. Additionally, we factorize the noise matrix based on feature and label representations, applying group sparsity constraints \cite{chartrand2013nonconvex} to model instance and label-dependent noise. The true label distribution space is essentially a lower-dimensional representation of the high-dimensional feature space \cite{rasiwasia2008scene}, sharing the same topological structure. To accurately reconstruct the true label distribution matrix, we employ graph regularization to align the topological structures of the input and output spaces. Finally, we use the Alternating Direction Method of Multipliers (ADMM) \cite{boyd2011distributed} for joint optimization, demonstrating that with a sufficient sample size, our method can recover the true labels and provide a generalization error bound. Our contributions are summarized as follows:

\begin{itemize}
	\item We introduce the concept of DN-ILDL, which accurately reflects real-world scenarios of inaccurate label distribution learning.
	\item We propose a method to handle Inaccurate Label Distribution Learning with Dependent Noise (ILDLDN) and validate its effectiveness on numerous real-world datasets.
	\item We present the range of noise recovery errors and establish generalization error bounds for the proposed method.
\end{itemize}

\section{ Related Work}
\textbf{Label Distribution Learning (LDL)}:
LDL introduces label distributions as a novel learning paradigm to quantify the relevance of each label, drawing significant interest from researchers. This section provides a concise review of LDL research. LDL methods are generally classified into three categories: problem transformation (PT), algorithm adaptation (AA), and specialized algorithm (SA). In PT, works like Geng \cite{geng2014multilabel} and Borchani et al. \cite{borchani2015survey} recast the LDL challenge as a single-label task using label probabilities as weights. AA methods modify traditional classifiers to meet LDL's unique needs, such as AA-kNN \cite{geng2016label}, which leverages neighbor distances to estimate label distributions. SA approaches often employ custom algorithms; for instance, LDL-SCL \cite{zheng2018label} improves prediction accuracy by utilizing local sample correlations, and Ren \cite{ren2019label} enhances model performance by learning both common and label-specific features simultaneously. LDL-LRR \cite{jia2021label} integrates a ranking loss function to better represent label ranking relationships. Although effective, these approaches typically assume precise label data, overlooking the common issue of annotation noise in real-world settings \cite{kou2023instance, xu2017incomplete}.

\textbf{Inaccurate Label Distribution Learning}:
The challenge of noise in LDL has received scant attention until recently. Kou \cite{kou2023inaccurate} pioneered the concept of learning from inaccurate label distributions, employing techniques like low-rank and sparse decomposition to correct label distributions affected by Gaussian noise. The idea of instance-dependent inaccurate LDL was introduced in \cite{kou2023instance}, acknowledging that noise may vary based on specific instances. More recent developments, such as GCIA \cite{he2024generative}, propose a generative approach using variational inference to improve LDL annotations by linking similar features to latent label distributions and modeling annotation errors through a confusion matrix.
Existing models often incorrectly assume that label noise is feature and label independent, a presumption rarely valid in practical applications. A more prevalent scenario involves dependent noise, where labels are influenced by both the labels and the instances. Next, we will define this problem more formally.

\section{The DI-ILDL Approach}
$\textbf{Preliminaries}$: Let \(\mathbf{X} \in \mathbb{R}^{n \times d}\) be the instance matrix, and let \(Y = \{y_1, y_2, \ldots, y_q\}\) represent the label space, where \(n\), \(q\), and \(d\) denote the numbers of instances, labels, and feature dimensions, respectively. The unknown ground-truth label distribution for instance \(\mathbf{X}\) is given by the matrix \(\mathbf{D} \in \mathbb{R}^{n \times q}\), where each row \(\mathbf{d}_i = [d_{\mathbf{x}_i}^{y_1}, \ldots, d_{\mathbf{x}_i}^{y_q}]^{\mathrm{T}}\) represents the label distribution vector for instance \(\mathbf{x}_i\). Here, \(d_{\mathbf{x}_i}^{y_l}\) indicates the label description degree of \(y_l\) for \(\mathbf{x}_i\). Each instance's supervised information must conform to the probability simplex, meaning \(\sum_{j=1}^q d_{\mathbf{x}_i}^{y_j} = 1\) for all \(i \in [n]\) and \(d_{\mathbf{x}_i}^{y_j} \geq 0\) for all \((i, j) \in [n] \times [q]\).

The corrupted label distribution matrix is \(\mathbf{\Omega} \in \mathfrak{D}^{n \times q}\), where \(\mathfrak{D}\) aligns with \(\mathbb{R}\) under noise. Our goal is to identify noisy labels in \(\mathbf{\Omega}\) and develop a decision function \(\mathfrak{G}: \mathbb{R}^{n \times d} \to \mathbb{R}^{n \times q}\) using the training set \(\{\mathbf{X}, \mathbf{\Omega}\}\) to closely replicate the true label distributions, ideally achieving \(\mathfrak{G}(\mathbf{X}_{i:}) \approx \mathbf{D}_{i:}\).
\subsection{Algorithm}
We aim to utilize the instance matrix \(\mathbf{X}\) and the assigned label distribution matrix \(\mathbf{\Omega}\) to train a novel ILDL model to predict true labels for previously unseen data. In practice, the annotated label matrix \(\mathbf{\Omega}\) includes noisy labels, which can be decomposed into a true label matrix and a noise label matrix. We use a linear regression model \cite{kou122024exploiting} for prediction and optimize the weight matrix \(\mathbf{W} \in \mathbb{R}^{d \times q}\) by minimizing the squared loss \cite{lee1996importance}. Recognizing the common assumption in multi-label learning that label spaces are correlated \cite{zhu2017multi, huang2012multi, sun2021global}, we assume a low-rank output space and employ the nuclear norm \cite{xu2017incomplete, sun2019partial} to capture this characteristic:
\begin{equation}
	\begin{aligned}
		& \min _{\mathbf{W}} \frac{1}{2}\|\mathbf{D}-\mathbf{X W}\|_F^2+\alpha\|\mathbf{W}\|_* ,
		\text { s.t. } \boldsymbol{\Omega}=\mathbf{D}+\mathbf{E},
	\end{aligned}
	\label{eq1}
\end{equation}
where \(\|\cdot\|_F\) and \(\|\cdot\|_*\) represent the Frobenius norm \cite{bottcher2008frobenius} and nuclear norm \cite{jaggi2010simple} respectively. Here, \(\alpha\) is a regularization parameter, \(\mathbf{D} \in \mathbb{R}^{n \times q}\) is the ground-truth label distribution matrix, and \(\mathbf{E} \in \mathbb{R}^{n \times q}\) is the noise label matrix.

Existing ILDL methods often assume that noise in labels is independent of features or labels. However, real-world evidence suggests that mislabeling is often closely related to specific instances and labels. To address this, we define the noise label matrix \(\mathbf{E}\) as the output from linear mappings of both features and labels, mathematically represented as \(\mathbf{E} = \mathbf{X P} + \mathbf{Y Q}\), where \(\mathbf{P} \in \mathbb{R}^{d \times q}\) and \(\mathbf{Q} \in \mathbb{R}^{q \times q}\) are coefficient matrices related to instances and labels, respectively. Given that noisy labels often arise from ambiguities in a limited number of cases, these coefficient matrices are inherently sparse, indicating that noise affects only certain key instances and labels. Previous research \cite{kou2023instance, xie2021partial} used the \(\ell_1\)-norm for inducing sparsity; however, the \(\ell_1\)-norm does not adequately capture noise dependent on specific instances or labels, leading to global sparsity instead of targeted sparsity. A more effective approach uses the \(\ell_{2,1}\)-norm \cite{ranjan2019tight}, which promotes group-level sparsity, making it suitable for identifying instances or labels often associated with noise. We employ the \(\ell_{2,1}\)-norm on matrices \(\mathbf{P}\) and \(\mathbf{Q}\) to ensure row sparsity that aligns with the structure of dependent noise. Thus, we model the dependent noise in a manner that specifically targets the most relevant instances and labels:
\begin{equation}
	\begin{aligned}
		& \min _{\mathbf{W}, \mathbf{P}, \mathbf{Q}}  \frac{1}{2}\|\mathbf{D}-\mathbf{X W}\|_F^2 
		+\alpha\|\mathbf{W}\|_*+\beta\|\mathbf{P}\|_{2,1}+\gamma\|\mathbf{Q}\|_{2,1}\\
		& \text { s.t. } \boldsymbol{\Omega}=\mathbf{D}+\mathbf{X P} + \mathbf{Y Q}, 
	\end{aligned}
	\label{eq2}
\end{equation}
where \(\beta\) and \(\gamma\) function as trade-off parameters that help balance the various components of our model. The \(\ell_{2,1}\)-norm, expressed as \(\|\mathbf{A}\|_{2,1} = \sum_i \sqrt{\sum_j \mathbf{A}_{ij}^2}\), promotes group sparsity among parameters. Our goal is to accurately recover the true label distribution matrix, preserve the intrinsic local relationships of the embedded feature data, and unveil its underlying manifold structure by aligning the topological structures of the output and feature spaces. Given that label distributions act as a low-dimensional representation of features, it is essential that the graph structure of the label space mirrors that of the feature space. Specifically, edge weights connecting samples in the feature space graph should align with those in the label distribution space graph, ensuring that high similarity between samples \(\mathbf{x}_i\) and \(\mathbf{x}_j\) in the feature space translates to a corresponding resemblance in the output space. This is modeled as \(\|\mathbf{S} - \tilde{\mathbf{S}}\|_F^2\), where \(\tilde{\mathbf{S}} = \sum_{i=1}^n \sum_{j=1}^n e^{-\frac{\|\mathbf{x}_i \mathbf{W} - \mathbf{x}_j \mathbf{W}\|_2^2}{\sigma}}\) represents the pairwise similarity matrix within the label space, and \(\mathbf{S}\), a pairwise similarity matrix in the feature space, measures neighbor proximity with \(\mathbf{S}_{ij} = \exp\left(-\frac{\|\mathbf{x}_i - \mathbf{x}_j\|_2^2}{\sigma}\right)\) for \(k\)-nearest neighbors, and zero otherwise. Here, \(\sigma\) serves as a hyperparameter to tune similarity magnitude. We simplify \(\tilde{\mathbf{S}}\) to \(\Phi(\mathbf{W}, \mathbf{X}, \sigma)\). By jointly optimizing Problem (\ref{eq2}) and the graph regularization term, we achieve the final formulation:
\begin{equation}
	\begin{aligned}
		& \min _{\mathbf{W}, \mathbf{P}, \mathbf{Q}}  \frac{1}{2}\|\mathbf{D}-\mathbf{X W}\|_F^2 
		+\alpha\|\mathbf{W}\|_*+\beta\|\mathbf{P}\|_{2,1}+\gamma\|\mathbf{Q}\|_{2,1} +\|\mathbf{S}-\mathbf{\tilde{S}}\|_F^2 \\
		& \text { s.t. } \boldsymbol{\Omega}=\mathbf{D}+\mathbf{X P} + \mathbf{Y Q}, \mathbf{\tilde{S}}= \Phi(\mathbf{W,X},\sigma).
	\end{aligned}
	\label{finall loss}
\end{equation}
\vspace{-3mm}
\subsection{  Optimizing using ADMM}
We employ the Alternating Direction Method of Multipliers (ADMM) \cite{boyd2011distributed} to address the optimization challenge presented in Eq. (\ref{finall loss}). For the sake of simplification, we incorporate an auxiliary matrix $ \mathbf{Z} = \mathbf{W}  \in  \mathbb{R}^{d \times q}$  to reformulate it equivalently. The augmented Lagrangian function is:

\begin{equation}
	\begin{aligned}
		\mathcal{L}=  &\frac{1}{2}\|\boldsymbol{\Omega}-\mathbf{X P} - \mathbf{Y Q}-\mathbf{X W}\|_F^2 
		+\alpha\|\mathbf{Z}\|_*+\beta\|\mathbf{P}\|_{2,1}+\gamma\|\mathbf{Q}\|_{2,1}\\ &+\|\mathbf{S}-\Phi(\mathbf{W,X},\sigma)\|_F^2 
		 +\left\langle\boldsymbol{\Gamma}, \mathbf{Z}-\mathbf{W}\right\rangle +\frac{\mu}{2}\|\mathbf{Z}-\mathbf{W}\|_F^2
	\end{aligned}
	\label{optimLOSS}
\end{equation}
where $\mu$ is a positive penalty parameter and $\boldsymbol{\Gamma} \in  \mathbb{R}^{d \times q}$ denotes the Lagrangian multipliers.  Eq. (\ref{optimLOSS}) can be solved by alternately optimizing three sub-problems as follow. The whole process is summarized in Algorithm \ref{A1}.

$\mathbf{1).Z}$-subproblem  is formulated as:
\begin{equation}
	\begin{aligned}
		& \min _{\mathbf{Z}} 
		 \alpha\|\mathbf{Z}\|_*+\frac{\mu}{2}\left\|\mathbf{Z}-\left(\mathbf{W}-\frac{\boldsymbol{\Gamma}}{\mu}\right)\right\|_F^2
	\end{aligned}
	\label{Z-SUB}
\end{equation}

Eq. (\ref{Z-SUB}) represents a nuclear norm minimization problem with a closed-form solution \cite{gu2014weighted}: $\mathbf{Z}=\mathbf{S}_{\frac{\alpha}{\mu}}\left(\mathbf{W}-\frac{\boldsymbol{\Gamma}}{\mu}\right)$, where $\mathbf{S}(\cdot)$ is the singular value thresholding function. This involves decomposing $\mathbf{W}-\frac{\boldsymbol{\Gamma}}{\mu}$ into its singular value decomposition (SVD) form $\mathbf{U} \boldsymbol{\Sigma} \mathbf{V}^{\top}$, followed by applying thresholding to derive $\mathbf{U} \boldsymbol{\Sigma}_{\alpha/\mu} \mathbf{V}^{\top}$, where each singular value is adjusted to $\Sigma_{\alpha/\mu, ii}=\max(0, \Sigma_{ii}-\alpha/\mu)$.

$\mathbf{2).W}$-subproblem  is formulated as:
\begin{equation}
	\begin{aligned}
			& \min _{\mathbf{W}}  \frac{1}{2}\|\boldsymbol{\Omega}-\mathbf{X P} - \mathbf{Y Q}-\mathbf{X W}\|_F^2 
		 +\|\mathbf{S}-\Phi(\mathbf{W,X},\sigma)\|_F^2 
		+\frac{\mu}{2}\left\|\mathbf{Z}-\left(\mathbf{W}-\frac{\boldsymbol{\Gamma}}{\mu}\right)\right\|_F^2
	\end{aligned}
	\label{W}
\end{equation}
which can be solved in a closed form $\mathbf{W} = (\mathbf{Q}+\mathbf{I})^{\top}  (\mathbf{P}  \mathbf{X} + (\mathbf{Q}+\mathbf{I})  \mathbf{W}  \mathbf{X}-\mathbf{\Omega})\mathbf{X}^{\top}  -\mu  (\mathbf{Z}-\mathbf{W}) -\mathbf{\Gamma} + \mathbf{W}_s
\label{W——slove}$,  where  $\mathbf{W}_s $ is provided in the appendix, and \( \mathbf{I}  \in \mathbb{R}^{q \times q} \) is the identity matrix.

$\mathbf{3).P}$-Subproblem and $\mathbf{Q}$-Subproblem  are  defined as follows:
\begin{equation}
	\begin{aligned}
		\min _{\mathbf{P}}   &\frac{1}{2}\|\boldsymbol{\Omega}-\mathbf{X P} - \mathbf{Y Q}-\mathbf{X W}\|_F^2 +\beta\|\mathbf{P}\|_{2,1}\\ 
	\end{aligned}
	\label{P_SUB}
\end{equation}

\begin{equation}
	\begin{aligned}
		\min _{\mathbf{Q}}   &\frac{1}{2}\|\boldsymbol{\Omega}-\mathbf{X P} - \mathbf{Y Q}-\mathbf{X W}\|_F^2 +\gamma\|\mathbf{Q}\|_{2,1}\\ 
	\end{aligned}
	\label{Q_SUB}
\end{equation}

To solve these, the gradient of Eq. (\ref{P_SUB}) is set to zero, yielding the solution: 
$$\mathbf{P} = (\mathbf{P} \mathbf{X} + (\mathbf{Q} + \mathbf{I}) \mathbf{W} \mathbf{X} - \boldsymbol{\Omega}) \mathbf{X}^\top + \beta  \text{diag}\left(\frac{1}{2} \|\mathbf{P}\|_2\right) \mathbf{P},$$ 
where diag($\mathbf{A}$) extracts diagonal elements from matrix $\mathbf{A}$, and $\mathbf{I}$ represents the identity matrix of appropriate size. Similarly, the solution for the $\mathbf{Q}$-Subproblem is given by:
$$\mathbf{Q} = (\mathbf{P}^\top \mathbf{X} + (\mathbf{Q} + \mathbf{I}) \mathbf{W} \mathbf{X} - \boldsymbol{\Omega}) \mathbf{X}^\top \mathbf{W}^\top + \gamma  \text{diag}\left(\frac{1}{2} \|\mathbf{Q}\|_2\right)  \mathbf{Q}.$$
Finally, the Lagrange multiplier matrix and penalty parameter $\mu$ are updated based on following
\begin{equation}
	\left\{
	\begin{array}{l}
	\begin{aligned}
	&\mathbf{\Gamma}=\mathbf{\Gamma}+\mu^k\left(\mathbf{Z}- \mathbf{W}\right) \\
	&\mu^{k+1}=\min \left(1.1 \mu, \mu_{\max }\right)
\end{aligned}
\end{array}\right.
\end{equation}
\begin{algorithm}
	\caption{Dependent Noise-based Inaccurate Label Distribution Learning (DN-ILDL)}
	\label{alg:DN-ILDL}
	\begin{algorithmic}[1]
		\REQUIRE Instance matrix $\mathbf{X} \in \mathbb{R}^{n \times d}$, noisy label matrix $\mathbf{\Omega} \in \mathbb{R}^{n \times q}$, and $\alpha, \beta, \gamma$, and $\sigma$
		\ENSURE Predicted true label distribution matrix $\mathbf{D}$
		\STATE Initialize weight matrices $\mathbf{W} \in \mathbb{R}^{d \times q}$, $\mathbf{P} \in \mathbb{R}^{d \times q}$, $\mathbf{Q} \in \mathbb{R}^{q \times q}$
		\STATE Define $\mathbf{D}$ as the true label distribution matrix with dimensions $n \times q$
		\STATE Calculate initial similarity matrix $\mathbf{S}$ for $\mathbf{X}$ using $k$-nearest neighbors and $\sigma$
		\REPEAT
		\STATE Update $\mathbf{W}, \mathbf{P}, \mathbf{Q}$ by minimizing Eq. (\ref{finall loss})
		\STATE Ensure each row of $\mathbf{D}$ sums to 1 and all elements are non-negative
		\STATE Recompute $\tilde{\mathbf{S}}$ using updated $\mathbf{W}$
		\UNTIL{convergence criterion is met}
		\RETURN $\mathbf{W},  \mathbf{P},  \mathbf{Q} $
	\end{algorithmic}
\label{A1}
\end{algorithm}
\vspace{-3mm}

\subsection{The Complexity Analysis}
The computational complexity of our algorithm is predominantly governed by operations such as matrix multiplications, singular value decomposition (SVD), and graph regularization. The core computations involve $\mathbf{XW}$, $\mathbf{XP}$, and $\mathbf{YQ}$, each with a complexity of $\mathcal{O}(n \times d \times q)$. Among these, the SVD step is notably the most demanding, essential for minimizing nuclear norms, and carries a complexity of $\mathcal{O}(\min(n^2 \times q, n \times q^2))$. Additional computational overheads include $\mathcal{O}(n \times q)$ for optimizing the $\ell_{2,1}$-norm and $\mathcal{O}(n^2 \times d)$ for implementing graph regularization. Collectively, the total computational complexity aggregates to $\mathcal{O}(n \times d \times q + \min(n^2 \times q, n \times q^2) + n^2 \times d)$.

\section{Theoretical Analysis}
Our theoretical analysis demonstrates that with a sufficiently large number of samples, the recovery error can be reduced to negligible levels. Below, we formalize this understanding through precise theorems.

\begin{theorem}
	Assume the actual noise matrices $\mathbf{E}^{*}$ depend on both instance and label characteristics, exhibiting group sparsity as indicated by sparsity levels $S_x$ and $S_y$, and group counts $G_x$ and $G_y$. With $\mathbf{W}$ fixed in Equation $(\mathcal{L})$, we consider $\mathbf{E}^{'} = \mathbf{\Omega} - \mathbf{XW}$ as the empirical observation of noise $\mathbf{E}^{*}$. Assuming that the discrepancy $\mathbf{E}^{'} - \mathbf{E}^{*}$ follows a sub-Gaussian distribution, our goal is to accurately derive the matrices $\mathbf{P}^*$ and $\mathbf{Q}^*$ from $\mathbf{E}^{'}$, akin to solving a group lasso problem. If $\beta \geq 2\epsilon(\sqrt{n} + \sqrt{6n\log n})$ and $\gamma \geq 2\epsilon(\sqrt{n} + \sqrt{6n\log q})$, where $\epsilon>0$ corresponds to the magnitude of observation error, and $G$ is the smaller of $\{G_x, G_y\}$ with dimensions $m_x = n$ and $m_y = q$, the bound for recovery error is:
	\[
	\|\mathbf{E}^{'} - \mathbf{E}^{*}\|_2 \leq \epsilon \sum_{i \in \{x, y\}} \sqrt{S_i} \left(\sqrt{\frac{1}{n}} + \sqrt{\frac{\log m_i}{n}}\right),
	\]
	with a probability exceeding $1 - 2/g^2$. This result suggests that our algorithm is likely to converge to the optimal solution, showing that the settings for $\beta$ and $\gamma$ do not depend on the sparsity level (noise rate). Provided these conditions are met and the sample size is sufficiently large, we can minimize the recovery error with high confidence, thereby eliminating the need for complex manual tuning of these parameters.
\end{theorem}
We further establish a generalization error bound for DI-ILDL. Defining the learned LDL function as $\mathcal{\xi}$, we describe the risk and empirical risk as $\mathcal{L}_{\mathcal{\kappa}}(\vartheta)$ and $\mathcal{L}_{\mathcal{\delta}}(\vartheta)$, respectively. The following theorem is then proved:

\begin{theorem}
	Let $\Xi$ be the family of functions for DI-ILDL. For any $\delta > 0$, with at least $1-\delta$ probability, for all $\mathcal{\xi} \in \Xi$, the following inequality holds:
	\[
	\mathcal{L}_{\mathcal{\kappa}}(\vartheta) \leq \mathcal{L}_{\mathcal{\delta}}(\vartheta) + \frac{4\sqrt{2}m(\sqrt{m}\epsilon + m\sigma)}{\sqrt{n}} + 6\sqrt{\frac{\log 2/\delta}{2n}}.
	\]
	This theorem articulates that the left-hand side represents the risk function, while the right-hand side sums the empirical risk, an upper bound of the Rademacher complexity, and a typically negligible third term. It sets an $\mathcal{O}(\nicefrac{m^2}{\sqrt{n}})$ generalization bound. Detailed proof is provided in the appendix.
\end{theorem}

\section{Experiments}
\subsection{Datasets and Evaluation Metrics}
\textbf{Datasets}: Our study utilizes 13 datasets\footnote{All datasets can be found at: \url{https://palm.seu.edu.cn/xgeng/index.htm\#codes}.} from various real-world domains to demonstrate the broad applicability and effectiveness of our method. The specifics of these datasets are detailed in Table \ref{dataset_detail}. The domains and specific datasets used are:
\begin{itemize}
	\item \textbf{Facial Beauty:} Includes M2B (ID: 1), SCUT-FBP (ID: 3), and Fbp5500 (ID: 4) \cite{ijcai2017p369}. These datasets are used to evaluate perceptions of facial beauty.
	\item \textbf{Facial Expression Analysis:} Utilizes RAF-ML (ID: 2) \cite{li2019blended} and SJAFFE (ID: 11) \cite{lyons1998coding}, which provide annotated data for facial expression analysis.
	\item \textbf{Sentiment Analysis:} Involves Flickr-ldl (ID: 5), Twitter-ldl (ID: 6) \cite{ijcai2017p456}, and Ren-Cecps (ID: 13) \cite{li2011creating}, offering data for sentiment analysis, including Chinese sentiment analysis.
	\item \textbf{Biological Data:} Includes Yeast experiments (ID: 7-8) \cite{geng2016label} and Human-Gene research (ID: 10), focusing on gene-disease interactions.
	\item \textbf{Nature Scenes:} Uses the Nature-scene dataset (ID: 12) \cite{geng2016label}, which contains multi-label images based on label distributions from rankings.
\end{itemize}
\textbf{Inaccurate Label Distribution Generation}:We generated synthetic noisy datasets to model instance- and label-dependent noise. First, using a defined noise rate $\pi$, we sampled instance flip rates $\mathbf{\varphi} \in \mathbb{R}^n$ from a truncated normal distribution $	\psi(\pi, 0.1^2, [0,1])$, and independently drew $\mathbf{\rho}_1 \in \mathbb{R}^{d \times q}$ and $\mathbf{\rho}_2 \in \mathbb{R}^{q \times q}$ from a standard normal distribution $\psi(0, 1^2)$. For each index $i$ in $[n]$, we computed instance- and label-dependent flip rates using $\mathbf{p}_i = \mathbf{\varphi}_i \times \operatorname{softmax}(\mathbf{x}_i \mathbf{\rho}_1 + \mathbf{d}_i \mathbf{\rho}_2)$. $\mathbf{p}_i$ is a probability vector summing to 1, matching the dimension of the features, from which we generate a corresponding selector vector $\mathbf{Sel}_i$ of equal size. Each element $\mathbf{Sel}_i(j)$ is set to 1 with a probability of $\mathbf{p}_i(j)$ and 0 otherwise. The label distribution for each instance, $\mathbf{d}_i$, is then updated by adding $\mathbf{Sel}_i$ and subsequently normalized to finalize the Inaccurate Label Distribution.

\textbf{Evaluation Metrics}:
We evaluate LDL algorithms using six metrics: five distance-based (Chebyshev, Clark, Kullback-Leibler, and Canberra) and two similarity-based (Cosine and Intersection). Formulas for these metrics are provided in the appendix. Lower values indicate better performance for distance-based metrics ($\downarrow$), while higher values indicate better performance for similarity-based metrics ($\uparrow$).
	\begin{table}[!t]\tiny\centering
	\centering\renewcommand{\arraystretch}{0.8}\tiny
	\setlength{\tabcolsep}{8mm}
	\caption{Details of the datasets.}
	\label{tab_dataset}
	\begin{tabular}{@{}lllllll@{}}
		\toprule
		Index & Dataset   & \#\textit{instances} & \#\textit{Features }& \#\textit{Labels} & \#\textit{Domain}  \\ 
		\midrule
		1     & M2B (M2B)  & 1,240     & 250       & 5    &  Images               \\ 
		2     & RAF-ML (RAF)   & 4,908     & 200       & 6     &Images\\
		3     & SCUT-FBP (SCU)  & 1,500     & 300       & 5   &Images                   \\ 
		4     & Fbp5500 (FBP) & 5,500     & 512       & 5   &Images                     \\ 
		5     & Flickr-ldl (Fli)  & 1,1150     & 200       & 8      &Images                  \\ 
		6     & Twitter-ldl (Twi) & 1,0040     & 200       & 8        &Images               \\ 
		7     & Yeast-cdc (Cdc)  & 2,465     & 24       & 15       &Biology              \\ 
		8     & Yeast-alpha (Alp) & 2,465     & 24       & 18  &Biology                    \\ 
		
	9     & SBU-3DFE (SBU)  & 2,500     & 243       & 6      &Images                 \\ 
		10    & Human-Gene(Gen) & 7,755     & 1869       & 5             &Biology          \\ 
		11    & SJAFFE  (SJA)   & 213      & 243      & 6        &Images                \\ 
		12    & Nature-scene   (Nat) & 2,000     & 294      & 9       &Images                 \\
		 13    & Ren-Cecps  (Ren) & 32420    & 100     & 8       & Text           \\
		\bottomrule
	\end{tabular}
\label{dataset_detail}
\end{table}

\subsection{Comparative Studies}
DI-ILDL was benchmarked against six established LDL methods and one ILDL approach, with hyperparameters configured according to their respective publications. For DI-ILDL, the trade-off parameters $\alpha$, $\beta$, and $\gamma$ were fine-tuned within the set $\{0.005, 0.01, 0.05, 0.1, 0.5, 1, 10\}$, $\sigma$ varied from 0.1 to 1, and $\pi$ remained fixed at 0.2. The competing methods are summarized as follows:
\begin{itemize}
	\item \textbf{LSR-LDL} \cite{kou2023instance}: Improves noise management by addressing inaccuracies specific to individual instances within label distributions.
	\item \textbf{LDL-LRR} \cite{9495131}: Integrates a ranking loss function with LDL to preserve the integrity of label rankings and enhance predictive performance.
	\item \textbf{LDLLDM} \cite{wang2021label}: Focuses on learning both global and local label distribution manifolds, emphasizing label interconnections and addressing incomplete label distribution learning.
	\item \textbf{EDL-LRL} \cite{jia2019facial}: Aims to capture local low-rank structures, enhancing exploitation of local label correlations.
	\item \textbf{IncomLDL} \cite{xu2017incomplete}: Utilizes trace-norm regularization and alternating direction methods, effectively leveraging low-rank label correlations.
	\item \textbf{LDLLC} \cite{zheng2018label}: Harnesses local label correlations to ensure closely aligned prediction distributions for similar instances.
	\item \textbf{LDL-SCL} \cite{jia2021label}: Considers the impact of local samples by encoding local label correlations, effectively learning label distribution.
\end{itemize}

\begin{figure*}[!h]
	\centering
	\subfloat[Intersection$\uparrow$]{
		\includegraphics[width=0.3\linewidth]{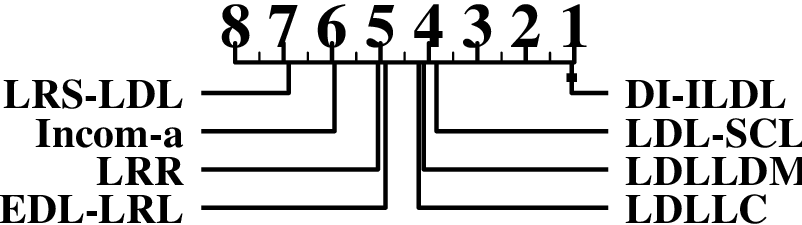}
	}
	\hfil
	\subfloat[Clark $\downarrow$]{
		\includegraphics[width=0.3\linewidth]{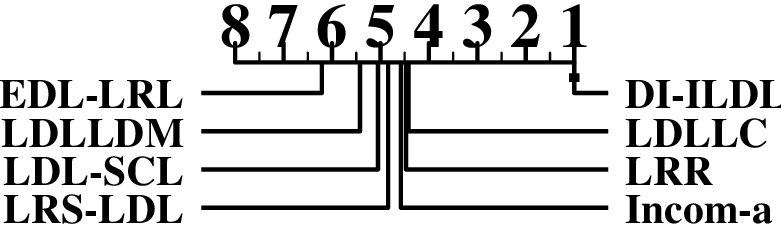}
	}
	\hfil
	\subfloat[Cosine$\uparrow$]{
		\includegraphics[width=0.3\linewidth]{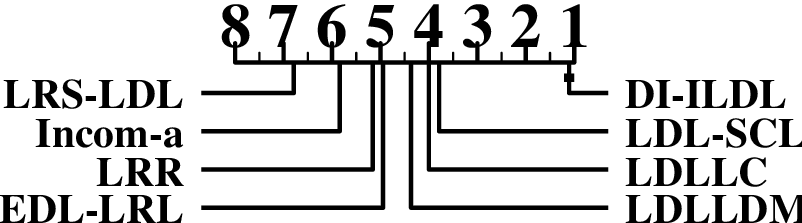}
	}
	\vfil
	\subfloat[Canberra $\downarrow$]{
		\includegraphics[width=0.3\linewidth]{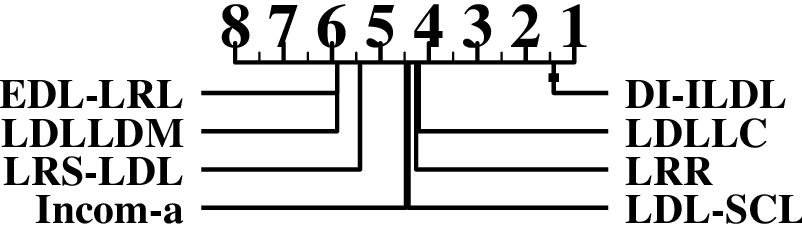}
	}
	\hfil
	\subfloat[Chebyshev$\downarrow$]{
		\includegraphics[width=0.3\linewidth]{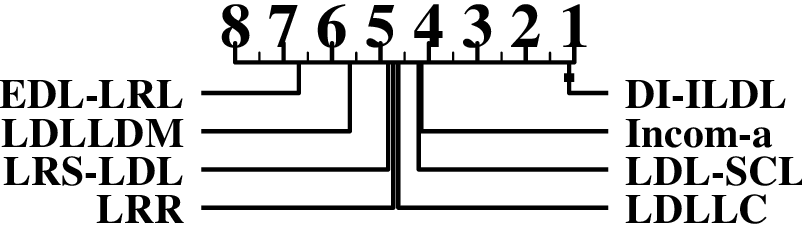}
	}
	\hfil
	\subfloat[KL$\downarrow$]{
		\includegraphics[width=0.3\linewidth]{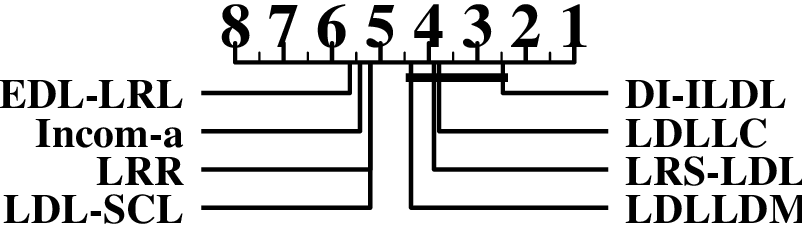}
	}
	
	\caption{CD diagrams of the comparing algorithms in terms of each evaluation criterion. For the tests, CD equals 2.3296 at 0.05 significance level.}
	\label{CD1}
\end{figure*}

\textbf{Results and Statistical Analysis:} Each method underwent ten runs on randomly partitioned data, with half used for training and the other half for testing. The results (mean$\pm$std.) are presented in Table \ref{zhushiyan1}, using Clark, Intersection, and KL metrics\footnote{The rest results mesured by other metric can be found in appendix.}, highlighting the best results in bold. Initially, the Friedman test \cite{demvsar2006statistical} evaluated the comparative performance of all methods (Table \ref{F-value}). At a confidence level of 0.05, the null hypothesis of equal performance for all algorithms was rejected. Subsequently, a Bonferroni-Dunn posthoc test was conducted, comparing the performance of DI-ILDL against other methods, using DI-ILDL as the control. Significant differences were noted when an algorithm's average rank differed by at least one critical difference (CD) \cite{demvsar2006statistical}, as illustrated in Figure \ref{CD1}. Algorithms with average ranks within one CD of DI-ILDL are connected by a thick line, indicating no significant performance difference. According to Table \ref{zhushiyan1}, DI-ILDL demonstrated exceptional performance, ranking first in 89.74\% of cases, and achieved the best mean performance across all metrics. Additional insights from Figure \ref{CD1} include:
\begin{itemize}
	\item DI-ILDL ranks first across all evaluation metrics, significantly outperforming 7 comparison algorithms on indicators other than KL distance, designed for learning from inaccurate label distributions based on dependent noise.
	\item DI-ILDL significantly outperforms EDL-LRR, Incom-a, LRR, and LDL-scl across all indicators, as these algorithms either only consider label correlation or focus solely on label ranking, disregarding label noise.
	\item Although DI-ILDL ranks first on the KL metric, it is not significantly different from LDLLC, LRS-LDL, and LDLLDM, as they consider label noise or label correlation.
\end{itemize}
\vspace{-3mm}

\begin{table}[]\tiny\centering\renewcommand{\arraystretch}{1}
	\setlength{\tabcolsep}{1.6mm}
	\caption{Predictive performance of each comparing approach (mean $\pm$ std) in terms of Clark distance$\downarrow$, Intersection similarity$\uparrow$ and KL distance$\downarrow$. $\uparrow(\downarrow)$ indicates the larger (smaller) the value, the better the performance. Best results are shown in boldface.}
	\begin{tabular}{ccccccccc}
		\hline\hline
		\multirow{2}{*}{Data sets} & \multicolumn{8}{c}{Clark distance $\downarrow$} \\ \cline{2-9} 
		 & DI-ILDL & LDLLC & Incom-a & LDL-SCL & LRR & LDLLDM & EDL-LRL & LRS-LDL \\ \hline
		alp & \textbf{0.2147±.0022} & 0.2240±.0051 & 0.2196±.0013 & 0.2172±.0026 & 0.2272±.0020 & 0.2211±.0010 & 0.2243±.0014 & 0.2333±.0040 \\
		cdc & \textbf{0.2192±.0015} & 0.2354±.0078 & 0.2324±.0060 & 0.2194±.0021 & 0.2458±.0063 & 0.2247±.0018 & 0.2317±.0011 & 0.2376±.0049 \\
		Fli & \textbf{0.2702±.0003} & 0.2745±.0127 & 0.2987±.0033 & 0.2712±.0001 & 0.3043±.0202 & 0.2712±.0084 & 0.2820±.0031 & 0.2743±.0011 \\
		Twi & \textbf{0.2937±.0003} & 0.2995±.0160 & 0.3350±.0042 & 0.3192±.0015 & 0.3586±.0201 & 0.3099±.0069 & 0.3077±.0191 & 0.3187±.0009 \\
		FBP & \textbf{1.4449±.0021} & 1.4471±.0091 & 1.4653±.0002 & 1.4768±.0325 & 1.5804±.0940 & 1.4724±.0007 & 1.5082±.0187 & 1.5086±.0003 \\
		Gen & \textbf{2.0755±.0022} & 2.1409±.0286 & 2.1278±.0125 & 2.1266±.0034 & 2.1301±.0139 & 2.1302±.0017 & 2.1046±.0055 & 2.1352±.0211 \\
		M2B & \textbf{1.6047±.0024} & 1.6498±.0121 & 1.6624±.0024 & 1.6748±.0151 & 1.6504±.0120 & 1.6648±.0061 & 1.6353±.0051 & 1.6894±.0120 \\
		Nat & \textbf{2.4259±.0180} & 2.4694±.0225 & 2.4805±.0186 & 2.4929±.0042 & 2.4672±.0058 & 2.4768±.0072 & 2.4852±.0085 & 2.4884±.0079 \\
		RAF & \textbf{1.3129±.0005} & 1.5663±.0076 & 13.0767±.4953 & 1.5828±.0071 & 1.5557±.0013 & 1.5769±.0074 & 1.5823±.0039 & 1.6122±.0051 \\
		Ren & \textbf{2.6072±.0008} & 2.6649±.0020 & 2.6584±.0025 & 2.6664±.0001 & 2.6647±.0005 & 2.6664±.0015 & 2.6658±.0001 & 2.6734±.0003 \\
		SBU & \textbf{0.4092±.0045} & 0.4130±.0089 & 0.4462±.0249 & 0.4120±.0011 & 0.4103±.0023 & 0.4111±.0045 & 0.4093±.0089 & 0.4144±.0016 \\
		SCU & \textbf{1.4863±.0024} & 1.4998±.0076 & 3.5421±.5205 & 1.4917±.0024 & 1.5013±.0052 & 1.4983±.0010 & 1.5033±.0023 & 1.4935±.0126 \\
		SJA & \textbf{0.4199±.0094} & 0.4357±.0248 & 0.4245±.0082 & 0.4251±.0089 & 0.4265±.0056 & 0.4370±.0036 & 0.4235±.0073 & 0.4323±.0257 \\ \hline
		\multicolumn{9}{c}{Intersection similarity $\uparrow$} \\ \hline
	
		alp & \textbf{0.9615±.0001} & 0.9595±.0011 & 0.9603±.0002 & 0.9609±.0004 & 0.9589±.0006 & 0.9601±.0001 & 0.9593±.0002 & 0.9577±.0007 \\
		cdc & \textbf{0.9569±.0003} & 0.9535±.0015 & 0.9538±.0020 & 0.9567±.0006 & 0.9515±.0011 & 0.9557±.0007 & 0.9538±.0002 & 0.9528±.0010 \\
		Fli & \textbf{0.9249±.0004} & 0.9165±.0031 & 0.9108±.0007 & 0.9156±.0001 & 0.9079±.0066 & 0.9171±.0017 & 0.9144±.0013 & 0.9145±.0005 \\
		Twi & \textbf{0.9096±.0001} & 0.9059±.0045 & 0.8970±.0009 & 0.8994±.0004 & 0.8889±.0060 & 0.9025±.0018 & 0.9042±.0056 & 0.8993±.0004 \\
		FBP & \textbf{0.5911±.0012} & 0.5867±.0113 & 0.5591±.0071 & 0.5419±.0447 & 0.5025±.0375 & 0.5456±.0012 & 0.5290±.0183 & 0.5010±.0010 \\
		Gen & \textbf{0.7833±.0006} & 0.7810±.0031 & 0.7827±.0014 & 0.7832±.0004 & 0.7829±.0019 & 0.7827±.0003 & 0.7858±.0004 & 0.7824±.0020 \\
		M2B & \textbf{0.4946±.0041} & 0.4477±.0136 & 0.4288±.0089 & 0.4137±.0182 & 0.4363±.0059 & 0.4283±.0073 & 0.4538±.0097 & 0.3896±.0089 \\
		Nat & \textbf{0.4037±.0026} & 0.3909±.0131 & 0.3791±.0066 & 0.3652±.0056 & 0.3954±.0018 & 0.3839±.0004 & 0.3701±.0095 & 0.3618±.0036 \\
		RAF & \textbf{0.5906±.0004} & 0.5527±.0030 & 0.0126±.0018 & 0.5264±.0028 & 0.5574±.0021 & 0.5342±.0027 & 0.5258±.0005 & 0.4934±.0001 \\
		Ren & \textbf{0.2857±.0004} & 0.2006±.0024 & 0.2184±.0055 & 0.1985±.0008 & 0.2012±.0019 & 0.1986±.0009 & 0.2000±.0002 & 0.1848±.0002 \\
		SBU & \textbf{0.8412±.0017} & 0.8391±.0037 & 0.8291±.0082 & 0.8390±.0007 & 0.8400±.0007 & 0.8392±.0018 & 0.8401±.0037 & 0.8379±.0005 \\
		SCU & \textbf{0.5264±.0018} & 0.5050±.0042 & 0.3107±.0296 & 0.5152±.0023 & 0.5021±.0017 & 0.5035±.0046 & 0.5008±.0015 & 0.5069±.0031 \\
		SJA & \textbf{0.8550±.0048} & 0.8447±.0088 & 0.8479±.0051 & 0.8482±.0033 & 0.8467±.0061 & 0.8408±.0006 & 0.8482±.0037 & 0.8455±.0131 \\ \hline
	\multicolumn{9}{c}{KL distance$\downarrow$} \\ \hline
		alp & \multicolumn{1}{l}{\textbf{0.0057±.0001}} & \multicolumn{1}{l}{0.0434±.0012} & \multicolumn{1}{l}{0.0059±.0001} & \multicolumn{1}{l}{0.0058±.0001} & \multicolumn{1}{l}{0.0063±.0001} & \multicolumn{1}{l}{0.0060±.0001} & \multicolumn{1}{l}{0.0062±.0001} & \multicolumn{1}{l}{0.0071±.0003} \\
		cdc & \multicolumn{1}{l}{0.0073±.0002} & \multicolumn{1}{l}{0.0501±.0017} & \multicolumn{1}{l}{0.0081±.0005} & \multicolumn{1}{l}{\textbf{0.0073±.0002}} & \multicolumn{1}{l}{0.0090±.0004} & \multicolumn{1}{l}{0.0075±.0001} & \multicolumn{1}{l}{0.0080±.0001} & \multicolumn{1}{l}{0.0091±.0005} \\
	
		Fli & \multicolumn{1}{l}{0.0236±.0001} & \multicolumn{1}{l}{0.1020±.0042} & \multicolumn{1}{l}{0.0303±.0004} & \multicolumn{1}{l}{0.0250±.0001} & \multicolumn{1}{l}{0.0305±.0038} & \multicolumn{1}{l}{0.0249±.0013} & \multicolumn{1}{l}{0.0264±.0006} & \multicolumn{1}{l}{\textbf{0.0228±.0002}} \\
		Twi & \multicolumn{1}{l}{0.0327±.0003} & \multicolumn{1}{l}{0.1187±.0066} & \multicolumn{1}{l}{0.0396±.0011} & \multicolumn{1}{l}{0.0369±.0003} & \multicolumn{1}{l}{0.0447±.0046} & \multicolumn{1}{l}{0.0347±.0011} & \multicolumn{1}{l}{0.0335±.0039} & \multicolumn{1}{l}{\textbf{0.0320±.0003}} \\
		FBP & \multicolumn{1}{l}{\textbf{0.5031±.0008}} & \multicolumn{1}{l}{0.5954±.0223} & \multicolumn{1}{l}{0.5822±.0100} & \multicolumn{1}{l}{0.6055±.1040} & \multicolumn{1}{l}{1.0622±.4218} & \multicolumn{1}{l}{0.5934±.0046} & \multicolumn{1}{l}{0.7731±.0722} & \multicolumn{1}{l}{4.2374±.0171} \\
		Gen & \multicolumn{1}{l}{0.2316±.0033} & \multicolumn{1}{l}{0.3778±.0083} & \multicolumn{1}{l}{0.2391±.0018} & \multicolumn{1}{l}{0.2386±.0008} & \multicolumn{1}{l}{0.2384±.0037} & \multicolumn{1}{l}{0.2399±.0001} & \multicolumn{1}{l}{0.2326±.0004} & \multicolumn{1}{l}{\textbf{0.2304±.0037}} \\
		M2B & \multicolumn{1}{l}{\textbf{0.8039±.0046}} & \multicolumn{1}{l}{0.9256±.0350} & \multicolumn{1}{l}{0.8726±.0131} & \multicolumn{1}{l}{0.9025±.0437} & \multicolumn{1}{l}{0.8676±.0033} & \multicolumn{1}{l}{0.8664±.0127} & \multicolumn{1}{l}{0.8240±.0161} & \multicolumn{1}{l}{1.7081±.0428} \\
		Nat & \multicolumn{1}{l}{\textbf{1.0148±.0135}} & \multicolumn{1}{l}{1.0750±.0327} & \multicolumn{1}{l}{1.1111±.0238} & \multicolumn{1}{l}{1.1463±.0236} & \multicolumn{1}{l}{1.0551±.0055} & \multicolumn{1}{l}{1.0830±.0032} & \multicolumn{1}{l}{1.1422±.0204} & \multicolumn{1}{l}{3.6774±.0219} \\
		RAF & \multicolumn{1}{l}{\textbf{0.5445±.0002}} & \multicolumn{1}{l}{0.6860±.0059} & \multicolumn{1}{l}{2.3730±.0483} & \multicolumn{1}{l}{0.6471±.0063} & \multicolumn{1}{l}{0.5758±.0028} & \multicolumn{1}{l}{0.6288±.0068} & \multicolumn{1}{l}{0.6491±.0015} & \multicolumn{1}{l}{4.2426±.0672} \\
		Ren & \multicolumn{1}{l}{\textbf{1.6071±.0019}} & \multicolumn{1}{l}{1.6929±.0107} & \multicolumn{1}{l}{1.6126±.0209} & \multicolumn{1}{l}{1.7007±.0043} & \multicolumn{1}{l}{1.6891±.0097} & \multicolumn{1}{l}{1.7014±.0045} & \multicolumn{1}{l}{1.6942±.0013} & \multicolumn{1}{l}{11.1500±.0023} \\
		SBU & \multicolumn{1}{l}{\textbf{0.0731±.0010}} & \multicolumn{1}{l}{0.2181±.0055} & \multicolumn{1}{l}{0.0955±.0080} & \multicolumn{1}{l}{0.0842±.0006} & \multicolumn{1}{l}{0.0833±.0006} & \multicolumn{1}{l}{0.0844±.0006} & \multicolumn{1}{l}{0.0833±.0033} & \multicolumn{1}{l}{0.0763±.0003} \\
		SCU & \multicolumn{1}{l}{\textbf{0.6661±.0040}} & \multicolumn{1}{l}{0.7982±.0151} & \multicolumn{1}{l}{1.3179±.1011} & \multicolumn{1}{l}{0.6759±.0072} & \multicolumn{1}{l}{0.7143±.0057} & \multicolumn{1}{l}{0.7273±.0105} & \multicolumn{1}{l}{0.7189±.0059} & \multicolumn{1}{l}{4.0594±.1373} \\
		SJA & \multicolumn{1}{l}{\textbf{0.0669±.0025}} & \multicolumn{1}{l}{0.1987±.0139} & \multicolumn{1}{l}{0.0720±.0036} & \multicolumn{1}{l}{0.0726±.0023} & \multicolumn{1}{l}{0.0737±.0046} & \multicolumn{1}{l}{0.0785±.0001} & \multicolumn{1}{l}{0.0717±.0029} & \multicolumn{1}{l}{0.0720±.0092} \\ \hline\hline
	\end{tabular}
\label{zhushiyan1}
\end{table}

\begin{table*}[!h]\caption{Summary of the Friedman statistics $F_F$ 
		in terms  of six evaluation metrics, as well as the critical value at a significance level of 0.05 (8 algorithms on 13 datasets). }
	\small
	\centering
	\renewcommand{\arraystretch}{1.1}
	\setlength{\tabcolsep}{0.9mm}
	\begin{tabular}{@{}l|ccccccc@{}}
		\hline
		Critical   Value ($\alpha=0.05$) & Evaluation metric & Chebyshev & Canberra & Cosine & Clark & Intersection & KL \\ \hline
		\multicolumn{1}{c|}{2.121 } & Friedman Statistics $F_F$ & 49.667 & 36.667 & 45 & 44 & 48.308 & 43.769 \\ \hline
	\end{tabular}
	\label{F-value}
\end{table*}

\vspace{-4mm}
\subsection{Further Analysis}

\begin{figure*}[!h]
	\centering
	\subfloat[Clark]{
		\includegraphics[scale=0.11]{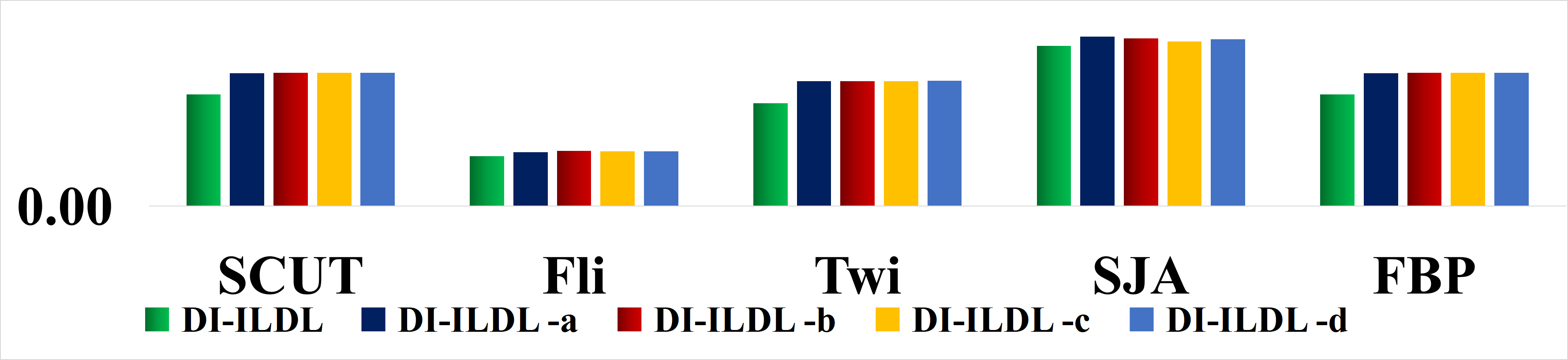}
	}
	\hfil
	\subfloat[KL]{
		\includegraphics[scale=0.11]{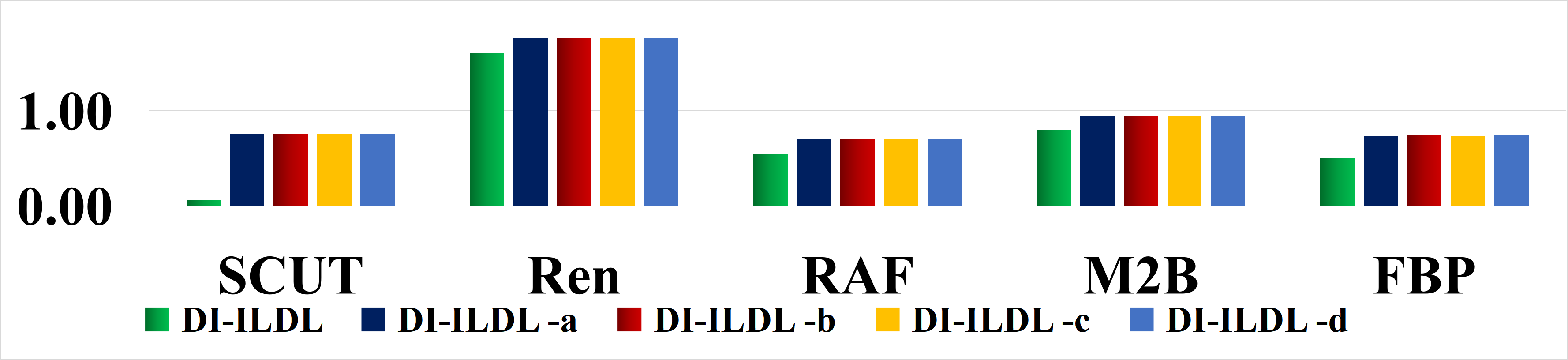}
	}
	\hfil
	
	\caption{ Ablation results on seven datasets  in terms of Clark $\downarrow$, KL $\downarrow$.}
	\label{xiaorongkeshihua}
\end{figure*}

\textbf{Ablation Study:} To rigorously evaluate the efficacy of our method in handling incorrectly labeled distributions with dependencies, we conducted an ablation study. This study involved sequentially removing the second, third, fourth, and graph regularization components from Eq. \ref{finall loss}, with each variant of the method designated as DI-ILDL-(a-d). The impacts of these modifications were assessed using Clark and KL divergence metrics, as depicted in Figure \ref{xiaorongkeshihua}. Additionally, the Wilcoxon signed-rank test was employed to analyze the statistical significance of performance differences between DI-ILDL and its variants, with the results documented in Table \ref{w-text}. Our findings are summarized as follows:
\begin{enumerate}
	\item \textbf{Label Correlation:} Incorporation of label correlation significantly improves the recovery of true labels and enhances prediction accuracy, highlighting its importance in the robustness of our method.
	\item \textbf{Noise Modeling:} The method's inclusion of group sparsity effectively addresses instance-dependent and label-dependent noise, thus efficiently managing dependency noise and enhancing label accuracy.
	\item \textbf{Graph Regularization:} The graph regularization component is crucial for aligning the topological structures of the output space with those of the input space, essential for accurate label recovery.
\end{enumerate}
These results confirm the critical contributions of each component to the overall effectiveness of our method, particularly in scenarios involving dependent noise in label distributions.

\begin{table}[!h]\renewcommand{\arraystretch}{1.2}
	\small
	\centering
	\setlength{\tabcolsep}{1.5mm}
	\caption{The results (Win/Tie/Loss[$p$-value]) of the Wilcoxon signed-rank tests for DI-ILDL against DI-ILDL-a, DI-ILDL-c, and DI-ILDL-d at a confidence level of 0.05.}
	\begin{tabular}{ccccccc}
		\hline\hline
		DI-ILDL\textit{\textbf{vs}.} & Chebyshev$\downarrow$ & Clark$\downarrow$ & Canberra$\downarrow$ & KL$\downarrow$ & Cosine$\uparrow$ & Intersection $\uparrow$ \\ \hline
		DI-ILDL-a & win{[}9.29e-05{]} & win{[}9.29e-05{]} & win{[}7.18e-05{]} & win{[}9.11e-05{]} & win{[}6.54e-03{]} & win{[}4.73e-03{]} \\
		DI-ILDL-b & win{[}4.73e-05{]} & win{[}4.73e-05{]} & win{[}1.39e-04{]} & win{[}2.15e-04{]} & win{[}4.21e-04{]} & win{[}2.14e-03{]} \\
		DI-ILDL-c & win{[}1.37e-04{]} & win{[}1.37e-04{]} & win{[}7.41e-05{]} & win{[}4.66e-05{]} & win{[}5.26e-04{]} & win{[}3.93e-04{]} \\
		DI-ILDL-d & win{[}4.21e-04{]} & win{[}4.21e-04{]} & win{[}3.79e-04{]} & win{[}4.25e-04{]} & win{[}6.29e-04{]} & win{[}3.62e-04{]} \\ \hline\hline
	\end{tabular}
\label{w-text}
\end{table}

\begin{figure}[!h]
	\centering
	\centering
	{
		\includegraphics[scale=0.27, clip]{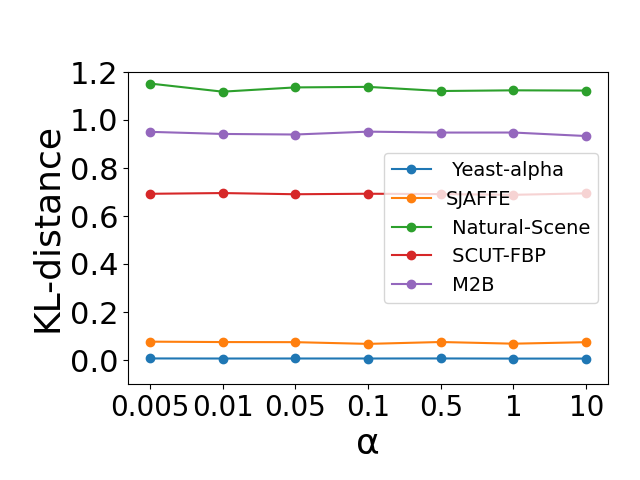}
	}
	\hfil
	{
		\includegraphics[scale=0.27, clip]{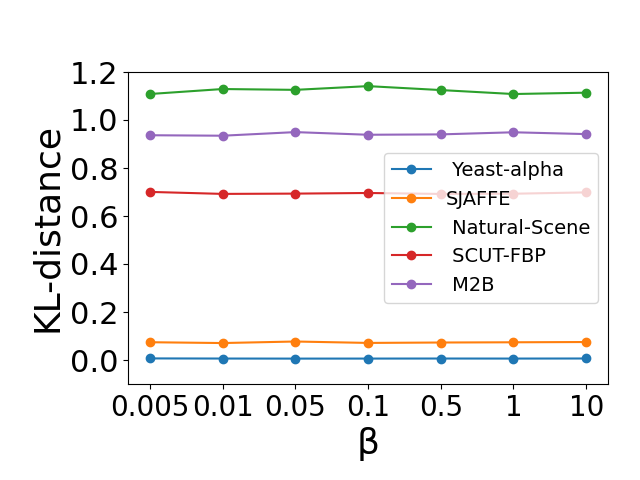}
	}
	\hfil
	{
		\includegraphics[scale=0.27, clip]{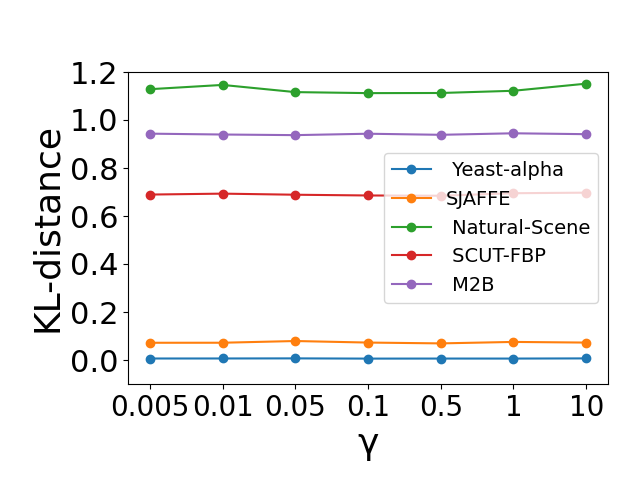}
	}
	
	\caption{The performance of DI-ILDL with $\alpha$, $\beta$ and $\gamma$ varying from $\left\{0.005, 0.01, 0.05, 0.1, 0.5, 1, 10\right\}$ in terms of KL-ditance  on five datasets.}
	\label{canshufenxi}
\end{figure}

\textbf{Parameter Sensitivity Analysis: }Fig. \ref{canshufenxi} illustrates the performance of the DI-ILDL algorithm across five datasets, evaluated using the KL-distance metric with parameters $\alpha$, $\beta$, and $\gamma$ adjusted within the range $\{0.005, 0.01, 0.05, 0.1, 0.5, 1, 10\}$. The graphs reveal a consistent pattern of minimal KL-distance for all parameter settings across each dataset, underscoring the robustness of DI-ILDL's performance against parameter variations. This consistency suggests that the algorithm operates with high stability and delivers uniformly strong performance across diverse settings, obviating the need for precise parameter tuning of $\alpha$, $\beta$, and $\gamma$.

\begin{figure}[!h]\centering
	\centering
	\centering
	\subfloat[]{
		\includegraphics[scale=0.21, clip]{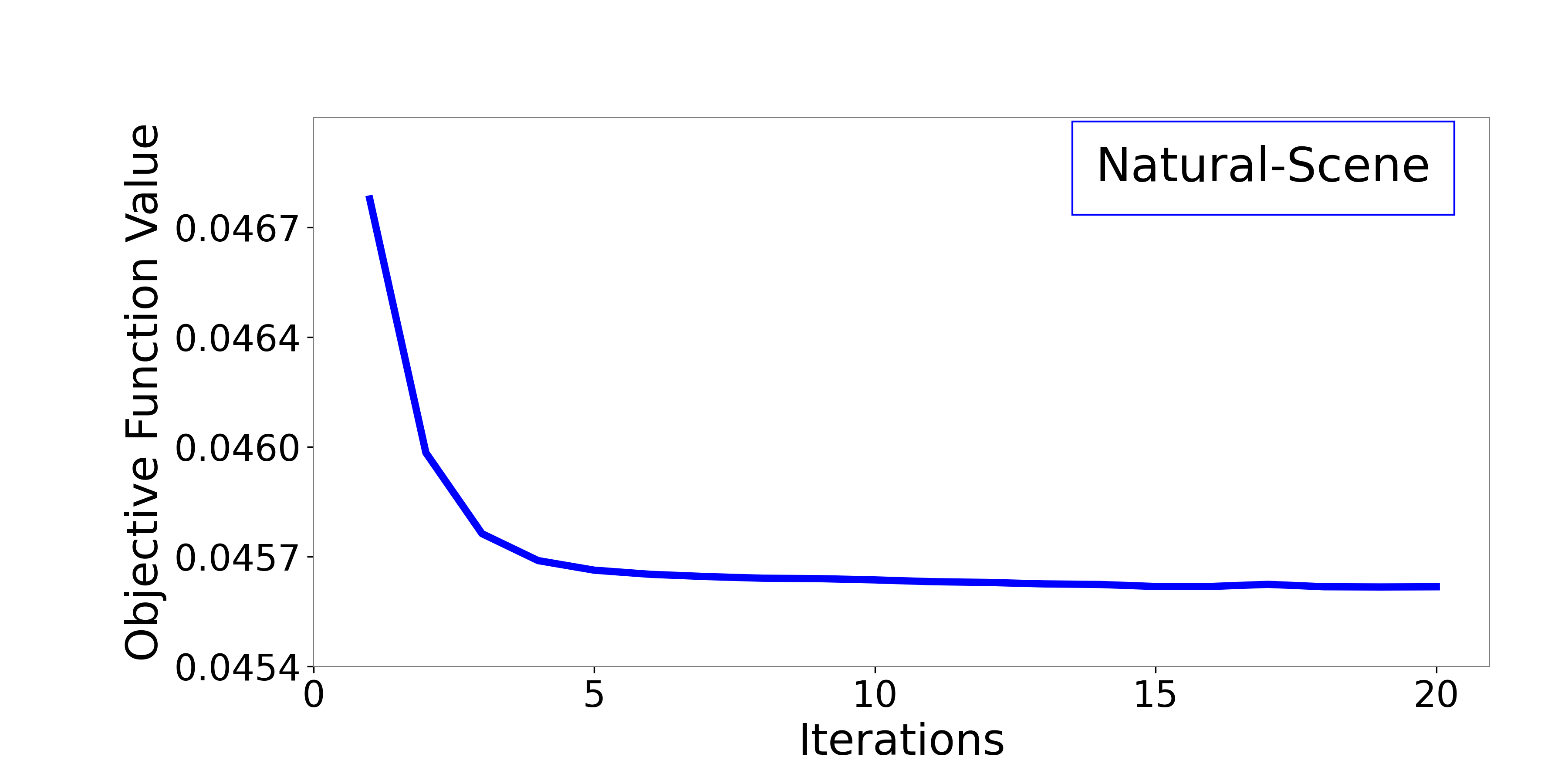}
	}
	\hfil
	\subfloat[]{
		\includegraphics[scale=0.21,  clip]{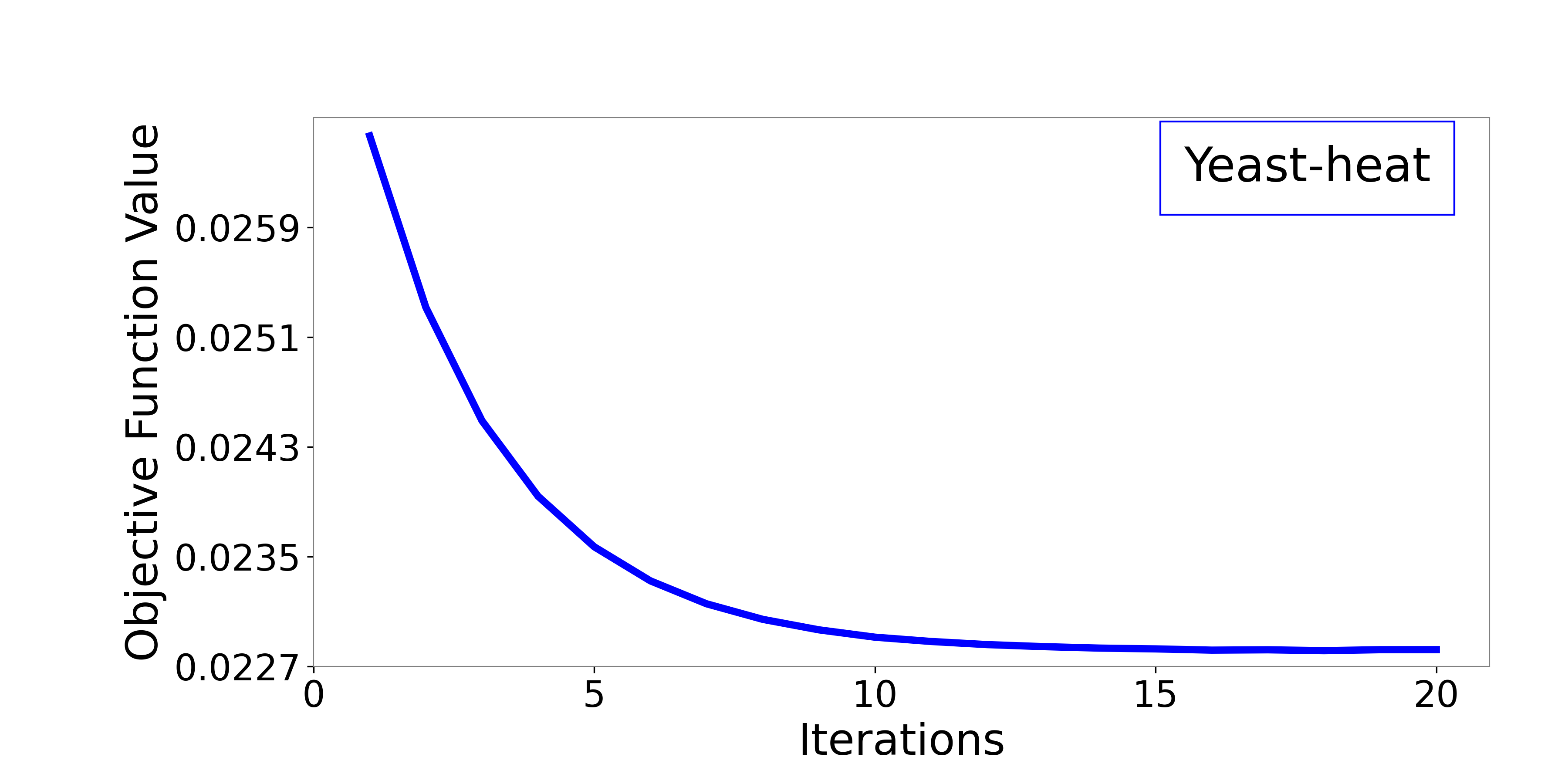}
	}
	
	\caption{Convergence of the objective functions of Eq. (\ref{finall loss}) with respect to thenumber of iterations on (a) Natural-Scene and (b) Yeast-heat.}
	\label{shoulianfenxi}
\end{figure}

\vspace{-3mm}
\textbf{Convergence Analysis:} Fig. \ref{shoulianfenxi} illustrates the convergence behavior of the objective functions for the Natural-Scene and Yeast-heat datasets over 20 iterations. Both graphs show a rapid decline in the objective values during the early iterations, followed by a quick stabilization. Specifically, the objective function for the Natural-Scene dataset stabilizes at approximately 0.0454 after about 10 iterations, while for the Yeast-heat dataset, it reaches a steady state around 0.0227 by the 15th iteration. This rapid convergence pattern demonstrates the model's efficiency, reaching near-optimal states early in the process and indicating that satisfactory results can be achieved with fewer iterations.
\vspace{-3mm}
\vspace{-3mm}
\section{Conclusion}
\vspace{-5mm}
In this study, we introduced the Dependent Noise-based Inaccurate Label Distribution Learning (DN-ILDL) approach, specifically designed to address the complexities associated with instance-dependent and label-dependent noise within label distributions. By leveraging linear mappings, group sparsity, and graph regularization, DN-ILDL not only reconstructs accurate label distributions but also effectively aligns the high-dimensional feature space with its corresponding lower-dimensional representations. We further established that with a sufficiently large sample size \( n \), DN-ILDL can precisely and reliably recover the true label distribution from its noisy observations and set robust generalization error bounds. Comprehensive evaluations across a variety of real-world datasets have confirmed that DN-ILDL proficiently handles the inherent challenges of ILDL-DN, demonstrating its broad applicability and effectiveness in practical scenarios.\\
\textbf{Limitations}: While the DN-ILDL approach has demonstrated considerable success in handling inaccurate label distributions influenced by instance-dependent and label-dependent noise, it exhibits limitations in scenarios involving imbalanced datasets.  We will address this  issue in the future.

\bibliographystyle{plain}
\bibliography{neurips_2024}


\end{document}